\documentclass{article}
\usepackage{ijcai25}

\usepackage{times}
\usepackage{soul}
\usepackage{url}
\usepackage[hidelinks]{hyperref}
\usepackage[utf8]{inputenc}
\usepackage[small]{caption}
\usepackage{graphicx}
\usepackage{amsmath}
\usepackage{amsthm}
\usepackage{booktabs}
\usepackage{algorithm}
\usepackage[noend]{algorithmic}
\usepackage[switch]{lineno}

\usepackage{xcolor}
\usepackage{amsmath}
\usepackage{xr}
\usepackage{amssymb}
\usepackage{eulervm}
\usepackage{paralist}
\usepackage[T1]{fontenc}
\usepackage{xspace}
\usepackage{bm}
\usepackage{svg}
\usepackage{eucal}

\newcommand{\wsts}{\textsf{PBS}\xspace}
\newcommand{\pbs}{\textsf{PBS}\xspace}
\newcommand{\bs}{\textsf{BS}\xspace}

\newcommand{\trt}{\textsf{TT}\xspace}
\newcommand{\dt}{\textsf{DT}\xspace}

\definecolor{myyellow}{HTML}{D6B656}
\definecolor{myred}{HTML}{b85450}
\definecolor{mygreen}{HTML}{82B366}

\newcommand{\ignore}[1]{}

\newtheorem{example}{Example}

\title{Portfolio Beam Search:\\
Diverse Transformer Decoding for Offline Reinforcement Learning Using
Financial Algorithmic
Approaches}

\author{
Dan Elbaz$^1$
\and
Oren Salzman$^2$\\
\affiliations
\emails
dan.elbaz@intel.com,
osalzman@technion.ac.il
}

\begin{document}

\maketitle

\begin{abstract}
Offline Reinforcement Learning (RL) algorithms learn a policy using a fixed training dataset, which is then deployed online to interact with the environment.
Transformers, a standard choice for modeling time-series data, are gaining popularity in offline RL. In this context, Beam Search (\bs), an approximate inference algorithm, is the go-to decoding method.
In offline RL, the restricted dataset induces uncertainty as the agent may encounter unfamiliar sequences of states and actions during execution that were not covered in the training data. 
In this context, \bs  lacks two important properties essential for offline RL:
It does not account for the aforementioned uncertainty, and its greedy left-right search approach often results in sequences with minimal variations, failing to explore potentially better alternatives.

To address these limitations, we propose Portfolio Beam Search (\pbs), a simple-yet-effective alternative to \bs that balances exploration and exploitation within a Transformer model during decoding.
We draw inspiration from financial economics and apply these principles to develop an uncertainty-aware diversification mechanism, which we integrate into a sequential decoding algorithm at inference time.
We empirically demonstrate the effectiveness of \pbs on the D4RL locomotion benchmark, where it achieves higher returns and significantly reduces outcome variability. 
\end{abstract}

\section{Introduction}
Reinforcement Learning (RL) is concerned with an agent learning how to take actions in an environment to maximize the total reward it obtains. An RL agent typically learns by trial and error, involving \textit{online} interaction with the environment~\cite{r:1}. However, such online interaction may incur undesirable costs, time or risks. This is  relevant to many fields such as healthcare, autonomous driving, and recommendation systems~\cite{r:63,r:75,r:76,r:79,r:77,r:78}. 

In contrast, \textit{offline RL} allows agents to avoid  undesirable online interactions by using pre-collected data (batch) from other agents or human demonstrations.  While this approach mitigates risks associated with online exploration, it also introduces significant challenges. In offline RL, there are no guarantees regarding the quality of the batch data, as it may have been collected by an agent following a suboptimal policy. Therefore, the goal for the learned policy is to outperform the policy that generated the data. Consequently, it is necessary to execute a different sequence of actions than those stored in the batch, leading to encounters with unfamiliar state-action sequences.

This situation introduces \emph{subjective} or \emph{epistemic} uncertainty---uncertainty arising from a lack of knowledge or statistical evidence due to the limited size of the training batch or due to its quality. This uncertainty can result in erroneous value estimations, increasing variability in the results, which decreases stability and reliability. This phenomenon, known as \emph{distributional shift}~\cite{r:23}, is one of the central challenges in offline RL.

\subsection*{Offline RL with Transformers}
Transformers are a deep-learning architecture originally proposed for natural language processing (NLP) tasks~\cite{r:57}. They  have demonstrated powerful expressive capabilities, leading to their adoption in various domains where processing sequential information is crucial~\cite{r:122,r:124}. 
Recently, Transformer models have been successfully applied to various RL benchmarks, such as language instructions for vision tasks~\cite{r:128}, the Atari games suite~\cite{r:129}, and continuous control locomotion tasks~\cite{r:130}.

While earlier studies employed traditional RL algorithms, merely substituting recurrent networks with Transformer architecture~\cite{r:131,r:132}, our work builds on a more recent approach wherein Transformers are used as trajectory models. 
Pioneered by Decision Transformer~(\dt)~\cite{r:4} and Trajectory Transformer~(\trt)~\cite{r:3}, these techniques utilize Transformers to learn dynamics and reward structures, framing offline RL as a conditional sequence-modeling problem.
With  trajectory models at our disposal, we can simulate potential future scenarios to formulate plans, a principle known as model-based planning~\cite{r:1}.

Our work targets real-world applications, particularly locomotion problems involving high-dimensional continuous domains. As a concrete example, consider the task of bipedal locomotion. This problem supports multiple solutions, as it can encompass various motion styles using different postures, stride patterns, or gaits. Such problems pose significant challenges for offline RL, as they are often sensitive to even minor variations, especially when such variations are not explicitly included in the offline data batch. 
Moreover, in such problems, the vast number of possible trajectories makes exhaustive model exploration impractical. In fact, similar to NLP tasks, the number of possible sequences grows exponentially with sequence length, making exact inference NP-hard.
Thus, designing algorithms that can efficiently explore these overwhelmingly large domains remains a significant challenge.


As finding an exact solution to these sequential-decision problems is typically intractable, approximate inference algorithms such as Beam Search (\bs) are typically used. \bs{} is a heuristic-search algorithm that explores the search space in a greedy left-right fashion retaining only the top~$b$ scoring candidates sequences. 
In offline RL, \bs is repurposed as a planning or decoding algorithm that integrates reward signals, evaluating candidate trajectories based on their cumulative return. 
Despite the success of this approach in some offline RL tasks, naively modifying \bs{} to to consider only the maximum predicted returns has several disadvantages that may be detremental in certain applications. Here, we highlight the two most critical ones:



\begin{enumerate}[(i)]
    \item \textbf{Ignoring the Distributional Shift}:  
    When used as a planning algorithm, \bs is not designed to be robust against the distributional shift typical in offline RL. Pursuing the maximal reward trajectory without accounting for uncertainty can result in suboptimal outcomes
        
    \item \textbf{Insufficient Diversity}:
    Greedy approaches, such as \bs{}, tend to follow the highest rewarding trajectories, which can result in limited exploration and a lack of diversity in decoded solutions.
    In locomotion problems, identifying various motion modalities, including suboptimal ones, can be advantageous.
    While a cluster of similar motions might fail if the predicted trajectories prove infeasible, a diverse portfolio of behaviors can adapt to new situations by employing alternative strategies.

\end{enumerate}

\subsection*{Overview and Contributions}
To address the shortcomings of \bs when decoding Transformers for offline RL, we propose Portfolio Beam Search (\wsts),  an uncertainty-aware strategy to decode a set of diverse sequences. 

Our decoding approach is inspired by the finance sector, where investors demand a reward for bearing risk when making investment decisions. This approach was formalized by Harry Markowitz~[\citeyear{r:6}] who introduced this risk-expected return relationship in the mean-variance model, for which he was awarded a Nobel Prize in economics. 
The mean-variance model addresses the portfolio-optimization problem (formally defined in Sec.~\ref{subsec:portfolio-opt}) by identifying the optimal allocation of wealth across different assets to achieve a diversified portfolio by taking into account both the mean (expected return) and the variance (risk). 

Building on portfolio theory, our primary technical contribution is the application of these principles to develop \pbs, a Transformer decoding algorithm for offline RL. 
\pbs mitigates the adverse effects of the distributional-shift by balancing the desire to maximize rewards with the risk associated with incorrect actions.  
In the context of offline RL, we treat each candidate trajectory as an asset. When selecting action sequences to explore, we consider our limited computation budget as a form of wealth to be allocated across these trajectories. To optimize this allocation, we solve a convex portfolio-optimization problem and explore trajectories with probabilities proportional to their allocated ``wealth''.

To demonstrate the effectiveness of \wsts, we present results from continuous control tasks using the widely adopted D4RL offline RL benchmark~\cite{r:9}. Our method consistently outperforms other Transformer-based offline RL techniques, while maintaining comparable memory requirements and a small run-time overhead. Additionally, and crucial for the offline RL setting, our algorithm exhibits significant stability, substantially reducing result variance compared to other Transformer decoding methods.

\section{Preliminaries}
\label{sec:preliminaries}
\subsection{Offline RL}
A \emph{Markov decision process} (MDP) is a tuple $\mathcal{M} = \{\mathcal{S}, \mathcal{A}, r, P, \rho_0, \gamma\}$
where $\mathcal{S}$ and $\mathcal{A}$ are sets of states and actions,
$r : \mathcal{S} \times \mathcal{A} \rightarrow \mathbb{R}$ is the reward function,
$P$ represents the system's dynamics, describing how the environment changes in response to the agent's actions. It is a conditional probability distribution of the form $P(s_{t+1}|s_t, a_t)$, which indicates the probability of transitioning to the next state~$s_{t+1}$ given the current state~$s_{t}$ and the applied action~$a_{t}$. 
$\rho_0$~defines the initial state distribution
and $\gamma \in (0, 1]$ is a scalar discount factor that penalizes future rewards. 
A policy~$\pi$ is  a probability function~$\pi(a_t|s_t)$, representing the probability of taking action~$a_t$ at state~$s_t$. 
Simply put, an agent makes a decision $a \in \mathcal{A}$ based on the current state $s_t \in \mathcal{S}$. The environment responds to the agent's action by transitioning to the next state~$s_{t+1} \in \mathcal{S}$ and providing a reward $r_t \in R$. The goal is to find an optimal policy that maximizes the expected sum of discounted rewards. 

In the \emph{offline RL} problem, we are provided with a static, pre-collected dataset containing transitions from trajectory rollouts $\mathcal{D} = \{(s_t, a_t, s_{t+1}, r_t)_i\}$, where~$i$ is the sample index of experiences. The actions are collected from some (potentially unknown) behavior policy~$\pi_B$ (for example,~$\pi_{B}$ can be human demonstrations, random robot explorations, or both) and the next states and rewards are determined by the dynamics. The goal is to use $\mathcal{D}$ to learn, in an offline phase, a policy~$\pi$ for the underlying, unknown MDP $\mathcal{M}$, to be executed in the environment an online phase.

\subsection{Transformers for Offline RL: a Sequence-modeling Approach}
\label{sec:alg_background}

\begin{figure*}[t]
\centering
\includegraphics[width=\textwidth]{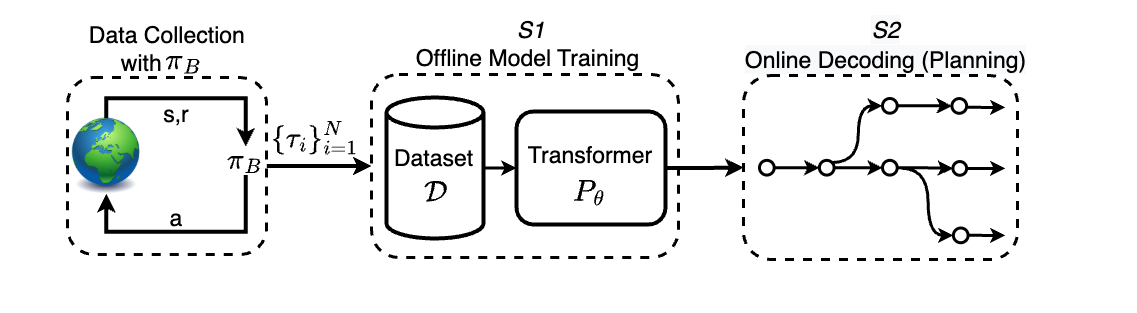} 
\vspace{-11.5mm}
\caption{
\textbf{Data collection:} The behavior policy $\pi_B$ collects the training batch $\mathcal{D}$.
\textbf{Model Training (S1, offline):} Using the previously collected dataset, $\mathcal{D}$, apply supervised learning to train a Transformer model, $P_{\theta}$, to estimate the distribution over trajectories.
\textbf{Decoding (S2, online):} During  inference, the decoding algorithm dynamically selects which trajectories to simulate at each step. The challenge lies in efficiently allocating computational resources to ensure that only the most promising trajectories are explored. Our work addresses this challenge, as formally defined in Sec. \ref{sec:method}.
\label{fig:top_level}
\vspace{-2.5mm}
}
\end{figure*}
Our work extends Trajectory Transformers~(\trt)~\cite{r:3} and Decision Transformers~(\dt)~\cite{r:4}, which redefine the offline RL problem as a conditional sequence-generation task, leveraging the Transformer architecture to learn the dynamics and rewards of sequential trajectories. Despite their simplicity, these techniques can achieve performance comparable to traditional RL algorithms that rely on complex optimization processes. 

Similar to other learning-based approaches, these methods divide the problem into two distinct stages: the learning stage for offline model training (\emph{S1}) and the inference stage for online decoding (\emph{S2}). Fig.~\ref{fig:top_level} illustrates this overall approach. Importantly, in this work, we focus on the decoding procedure, \emph{S2}, depicted in the right-hand side of Fig.~\ref{fig:top_level}.


\subsubsection{Setting \& Notation}
A transition is a tuple of state, action, and scalar reward, and is further augmented with an estimate of the discounted reward-to-go~$(R_t=\sum_{t' = t}^H\gamma^{t'-t}r_{t'})$.
Each trajectory in $\mathcal{D}$ consists of $H$ transitions. 
We denote a trajectory as  $\mathbf{T}$ and the~$t^{th}$ transition of  $\mathbf{T}$ as $\bm{\tau_t} = (s_t, a_t, r_t, R_t)$.
Assuming $K$-dimensional states and $M$-dimensional actions, each dimension is discretized independently. Thus, the dataset used to train the model consists of trajectories in the following form:

$$
\mathbf{T}
= (\ldots, 
s_t^1, s_t^2,\ldots, s_t^K, 
a_t^1, a_t^2,\ldots,a_t^M,  
r_t, R_t,
s_{t+1}^1
\ldots).
$$
In this representation~$\mathbf{T}$ is a sequence of length~$H \cdot (K +M +2)$.
For notational convenience, we represent elements of~$\mathbf{T}$ as a sequence of transitions $\mathbf{T} := (\bm{\tau_1}, \bm{\tau_2}, \ldots, \bm{\tau_H})$ and index the transitions by $t$, where $t$ is a natural number.
We treat these transitions as our vocabulary.
Accordingly, $\mathbf{T_{<t}}$ denotes a trajectory from the first time step up to transition $t-1$. 
Finally, $x$ is the initial state, sampled from~$\rho_0$ (the initial state distribution of the MDP~$\mathcal{M}$). 

\subsubsection{Offline Model Training (\textbf{\emph{S1})}}
As we will employ the same trained model from the \trt{} original work, we  briefly discuss the offline training stage, \emph{S1}. For a more comprehensive explanation of the architecture and the dataset creation process, we refer the reader to the original paper~\cite{r:3}

In this stage, a Transformer model $P_{\theta}$ (parametrized via the network's weights $\theta$) is trained using supervised learning on a given dataset\footnote{In our specific setting, $\mathcal{D}$ is the obtained from the MuJoCo physics simulator~\cite{r:127}.} $\mathcal{D}$ to predict the probability distribution over an output space of possible trajectories. To train the Transformer, we maximize the conditional log-likelihood on trajectories from the offline dataset.
This way, the Transformer model, learns to predict the probability distribution over an output space of possible trajectories given an initial state $x$.

Next, we use this model to generate trajectories token-by-token using a particular decoding strategy.

\subsubsection{\textbf{Online Decoding (\emph{S2})}}\label{subsec:decoding}
We begin with a refresher on \bs{} applied to offline RL, presenting it as a meta-algorithm.
Although this presentation is a bit unconventional and requires some overhead, it will significantly simplify the presentation of our planning algorithm, \wsts{}, as we will see in Sec.~\ref{sec:method}. 
Accordingly, we present in Alg.~\ref{alg:beam_search} \bs{} as an algorithm that receives as an input two functions: a scoring function {\tt \textbf{score}} and a pruning function {\tt \textbf{prune}} and returns a set of $b$ approximated \textit{best} trajectories (we will define \textit{best} shortly). 

More specifically, \bs{} maintains a set containing up to $b$ trajectories, collectively forming the current beam, $\mathcal{B}_t$. Initially, $\mathcal{B}_0$ includes only the start state $x$~(Line~\ref{line:init}). At each subsequent time step $t > 0$ within the planning horizon, \bs{} expands the trajectories in the current beam ($ \mathbf{T_{<t}} \in \mathcal{B}_{t-1}$) to generate new candidates and stores them in $\mathcal{C}$~(Line~\ref{line:expand}). Each of these candidates is generated by extending the existing trajectories in the current beam with potential transitions~$\mathbf{T_{<t}} \circ \bm{\tau_t}$ (here, $\circ$ denotes concatenation), where the transitions are sampled from the Transformer model~$P_{\theta}$ by conditioning on the trajectories from the current beam.
The candidate sequences in $\mathcal{C}$ ($\vert \mathcal{C}\vert \geq \vert \mathcal{B}_t \vert$) are then evaluated using the {\tt \textbf{score}} function~(Line~\ref{line:score}), which assigns a score $w_n$ to each candidate trajectory, $\mathbf{c_n} \in \mathcal{C}$. Subsequently, these scores serve as inputs to the {\tt \textbf{prune}} function, which determines which $b$ sequences to retain for the next iteration~(Line~\ref{line:prune}), typically selecting those with the highest scores.


\begin{algorithm}[t]
\caption{Beam Search}
\label{alg:beam_search}
\textbf{Input}: start state $x$, scoring function {\tt \textbf{score()}}, pruning function {\tt \textbf{prune()}}, Transformer model~$P_{\theta}$ \\
\textbf{Parameters}: beam width $b$, planning horizon $H$\\
\textbf{Output}: Approx. set of $b$ best trajectories.

\begin{algorithmic} [1] 
\STATE $\mathcal{B}_0 \gets \{x\} $ \label{line:init}
\FOR{$t \gets 1$ to $H$}
\STATE $\mathcal{C} \gets \{\} $
\FORALL{$ \mathbf{T_{<t}} \in \mathcal{B}_{t-1}$}
\STATE $\mathcal{C}\leftarrow \{ \mathbf{T_{<t}} \circ \bm{\tau_t} \mid \bm{\tau_t} \in P_{\theta}(\bm{\tau_t}| \mathbf{T_{<t}}) \}$ \label{line:expand}
\ENDFOR

\STATE $\mathbf{w} \gets {\texttt{\textbf{score}}}(\mathcal{C}, P_{\theta})$ \label{line:score}
\STATE $\mathcal{B}_t \gets {\texttt{\textbf{prune}}}(\mathcal{C}, b,\mathbf{w})$ \label{line:prune}
\ENDFOR
\RETURN $\mathcal{B}_t$

\end{algorithmic}
\end{algorithm}

\subsubsection{\textbf{\bs{} as a Planner}}\label{subsec:decoding}
When using Transformers for NLP, the overarching objective is to identify the most probable sequence under the model during inference, a process known as Maximum a Posteriori (MAP) decoding. 
In this case, \bs{} scores candidate sequences ,$\mathbf{c_n} \in \mathcal{C}$, in proportion to their probability, selecting the $b$ most likely sequences from candidates.
When this decoding scheme is applied to locomotion or control problems, the decoded sequence mirrors the demonstrations in the data batch, a technique commonly referred to as imitation learning or behavioral cloning~\cite{r:138}.

However, in offline RL, the quality of the demonstrations in the batch data isn't guaranteed, as it might have been gathered by an agent using a suboptimal policy. Therefore, the objective for the learned policy is to surpass the performance of the policy that generated the data. Thus, the log probabilities of transitions are replaced by the log probability of the predicted reward signal. Consequently, in Alg.~\ref{alg:beam_search}, the {\tt \textbf{score}} function (Line~\ref{line:score}) is modified to rank candidate trajectories in~$\mathcal{C}$, based on the expected cumulative reward plus the reward-to-go estimate (while this estimate reflects the return from the behavior policy $\pi_B$, it serves as a sufficient heuristic to complement the cumulative rewards).
In this case, the pruning function {\tt \textbf{prune}}~(Line~\ref{line:prune}) remains unchanged and selects the trajectories with the highest scores.

\subsection{Portfolio Optimization}
\label{subsec:portfolio-opt}
In finance, a \emph{portfolio} is a combination of financial assets, each associated with its individual expected return and risk. To construct a portfolio, given $N$ different assets and a given amount of wealth, we need to decide the allocation of wealth to each asset. This involves determining a weight vector for the $N$ assets, denoted as $\mathbf{w} = (w_1, w_2, \ldots, w_N)^T$, where all the elements in $\mathbf{w}$ are positive and their sum equals one.

The \emph{portfolio-optimization problem} (Eq.~\eqref{eq:portfolio_opt}) involves selecting the best portfolio from the set of all possible portfolios, according to a specified objective. 
The mean-variance model~\cite{r:6}, addresses the portfolio-optimization problem by considering the mean (term~(1)~in~Eq.~\eqref{eq:portfolio_opt}) and the variance and correlation between assets in a portfolio~(term~(2)~in~Eq.~\eqref{eq:portfolio_opt}). This approach balances expected reward and risk, enabling investors to maximize returns while managing risk. Additionally, it emphasizes the importance of diversifying investments to achieve a principled method of managing risk and return. 
Specifically, given some risk-aversion parameter~$\delta$,
assets mean vector~$\bm{\mu} \in \mathbb{R}^{N \times 1}$ and covariance matrix ~$\bm{\Sigma} \in \mathbb{R}^{N \times N}$,
solving the portfolio-optimization problem is equivalent to computing a weight vector $w$ dictating the relative amount to invest in each asset in the following optimization problem:
\begin{equation}
\label{eq:portfolio_opt}
\begin{aligned}
\underset{\mathbf{w}}{\text{Maximize}} \quad & \underbrace{ \mathbf{w}^T \bm{\mu}}_{(1)} - \underbrace{{\delta}\mathbf{w}^T \bm{\Sigma} \mathbf{w}}_{(2)} + \underbrace{\alpha \cdot \mathbf{w}^T \mathbf{w}}_{(3)},  \\
\text{subject to} \quad & \sum_{i=1}^N w_i=1 \quad \text{and} \quad w_i \geq 0.
\end{aligned}
\end{equation}

Mean-variance optimization frequently results in numerous negligible weights, leading to the exclusion of most assets from the portfolio. While this outcome is expected, it can be problematic when a specific number of assets are required. Consequently, the number of negligible weights is often reduced by incorporating a regularization term, $\alpha \cdot w^T w$, into the objective function~(term~(3)~in~Eq.~\eqref{eq:portfolio_opt}). 
We note that the overall problem is convex, which enables efficient resolution through the use of convex-optimization solvers.

\section{Method: Portfolio Beam Search (\wsts)}
\label{sec:method}

\begin{quote}
    \emph{There is nothing wrong with a `know nothing' investor who realizes it. The problem is when you are a `know nothing' investor, but you think you know something.}
\end{quote}
This quote, associated to Warren Buffett~\cite{r:61} 
aligns with core idea of our decoding method, Portfolio Beam Search~(\wsts), for Transformer decoding in offline RL. 

Similar to other sampling-based variants of \bs, \wsts is a best-first search method that retains at most $b$ trajectories at each time step. However, it differs in its definition of `best'. \wsts determines which trajectories to retain by considering both the expectation and the uncertainty due to distributional shift. Additionally, it assesses the similarity between candidate trajectories to promote diversity in the decoded solutions.
To achieve this, \wsts strategically allocates computational resources by solving a portfolio optimization problem. However, unlike standard portfolio optimization, in our context, the budget represents computational effort rather than monetary value, and the assets are the candidate trajectories.

To introduce \wsts, we revisit Alg.~\ref{alg:beam_search} which presented \bs as a meta-algorithm. Recall that this meta-algorithm incorporates two  functions: {\tt \textbf{score}} and {\tt \textbf{prune}} and outputs a set of $b$ approximated \textit{best} trajectories. We now describe \pbs via a description of these two functions.

\subsection{{\tt \textbf{score()}}} 
The \texttt{\textbf{score()}} function (Alg.\ref{alg:beam_search}, Line~\ref{line:score}), is the core of~\pbs. It assigns a relative score to each of the~$N$~candidate trajectories available in~$\mathcal{C}$ at time step~$t$. Specifically, we assign a relative score, $w_n$, to each candidate trajectory, $\mathbf{c_n}$, by solving a portfolio-optimization problem~(Eq.~\eqref{eq:portfolio_opt}) that encompasses all candidate trajectories~$\mathcal{C}$ (where $N = \vert \mathcal{C} \vert \geq b$). The solution to this optimization problem, represented by $ w_1, \ldots, w_N$, determines the proportion of computational time allocated to exploring each candidate trajectory.

To solve this portfolio-optimization problem, we need to construct two key components: a vector of expected returns for all candidate trajectories, denoted as $\bm{\mu}$, and a covariance matrix between these trajectories, denoted as $\bm{\Sigma}$, as outlined in terms (1) and (2) of Eq.~\eqref{eq:portfolio_opt}. This subsection details the methodology for constructing these components. Additionally, we require two hyperparameters: the regularization coefficient ($\alpha$) and the risk-aversion parameter ($\delta$), which will be further discussed in the experiments section.

Before we describe how to construct $\bm{\mu}$ and $\bm{\Sigma}$ for the candidate trajectories in $\mathcal{C}$, we begin with a preliminary computation to determine the mean and variance of the reward (or the reward-to-go) of a single transition $\bm{\tau}$.

\subsubsection{Single Transition Evaluation} 
\label{subsec:mean_cov}

Recall that a transition is represented as a tuple $\bm{\tau_t} = (s_t, a_t, r_t, R_t)$.
We sample full transitions according to the Transformer model log-probabilities (Alg. \ref{alg:beam_search}, Line~\ref{line:expand}).
However, to evaluate a transition, we consider the mean and variance of either the reward ($r_t$) or the reward-to-go ($R_t$), as higher values for these metrics are indicative of potentially better transitions. We do not consider the numerical values of actions or states, as higher values in these contexts do not correspond to better actions or states.

Accordingly, as a probabilistic classifier, the Transformer's softmax layer defines a multinomial probability distribution. Assuming a vocabulary $\mathcal{V}$, the network’s output layer consists of logits over this vocabulary. Specifically, for a discrete output variable $r_t$ (which, with a slight abuse of notation, represents a reward or reward-to-go estimate), the outputs correspond to a mapping: $r_t^i \rightarrow p_i$ for $i \in {1, \ldots, \vert \mathcal{V} \vert}$. Consequently, the mean $E[r_t]$ and variance $\mathrm{Var}[r_t]$ of the output~$r_t$ can be calculated from the Transformer's softmax layer as:
\begin{equation}
\label{eq:single_reward}
\begin{aligned}
 \quad & E[r_t] = \sum _{i=1}^{\mathcal{V}}p_{i}v_t^{i},  \\
\quad & \mathrm{Var}[r_t] = \sum_{i=1}^{\mathcal{V}}p_{i}\cdot (v_t^{i}-E[r_t] )^{2}. 
\end{aligned}
\end{equation}
Next, we utilize this single transition computation to evaluate candidate trajectories which are defined as a sequence of transitions $\mathbf{c_n} = (\bm{\tau_1}, \bm{\tau_2}, \ldots, \bm{\tau_t})$.

\subsubsection{{Trajectory Mean}} 
\label{subsec:mean_vector}
\label{subsec:mean_vector}
The mean vector, denoted as $\bm{\mu}$, is an $N$-dimensional vector where each entry $\mu_n$ corresponds to one of the $N$ candidate trajectories $\mathbf{c_n \in \mathcal{C}}$. To construct this vector, we calculate each entry $\mu_n$ using the previously described mean calculation (Eq.~\eqref{eq:single_reward}), applied to each transition within the~$n^{\text{th}}$ trajectory. This process involves accumulating the expected rewards for every transition in the trajectory and adding the reward-to-go estimate of the final transition as follows:
\begin{equation}
    \label{eq:mean}
    \mu_n =\sum_{t=1}^{H-1} \gamma^t \mathrm{E[r_t^n]}+ \gamma^T \mathrm{E[\hat{R}_{T}^n]}.
\end{equation}
Note that the computation of the expected returns vector, as presented in Eq.~\eqref{eq:mean}, follows the same methodology outlined in the original Trajectory Transformer and its subsequent papers. We include it here for completeness.

\subsubsection{Trajectories' Covariance}
\label{subsec:cov_matrix}

Recall (Sec.~\ref{sec:alg_background} and Eq.~\eqref{eq:portfolio_opt}) that the covariance matrix~$\bm{\Sigma}$ captures the relationships between different assets in a portfolio. 
It is standard practice to decompose the covariance matrix as: 
\begin{equation}
    \label{eq:decomposition}
    \bm{\Sigma} = \mathbf{U} \cdot \mathbf{S} \cdot \mathbf{U}.
\end{equation}
Here,~$\mathbf{U}$ is a diagonal matrix of standard deviations, often referred to as the risk or standard-deviation matrix, $\mathbf{S}$ is the correlation or similarity matrix and $\cdot$ represents matrix multiplication.
Next, we detail how~$\mathbf{U}$ and $\mathbf{S}$ are constructed in our specific domain.

\emph{Standard-Deviation Matrix ($\mathbf{U}$):} In the portfolio-optimization setting, each element along $\mathbf{U}$'s diagonal represents the risk or volatility of the corresponding asset. In our context, these elements correspond to the risk associated with candidate trajectories due to distributional shift.
We begin by noting that the variance in the Transformer’s output (Eq.~\ref{eq:single_reward}) measures predictive uncertainty, which, in turn, is a proxy for the distributional shift. This observation is supported by Desai et al.~[\citeyear{r:45}], who demonstrated that Transformer-based models exhibit well-calibrated performance when trained with temperature scaling~\cite{r:46}, a technique we also employ. 

Now, predicting the variance of the expected returns~$\sigma_n^2$, of the~$n^{\text{th}}$ candidate trajectory $\mathbf{c_n}$, necessitates careful consideration. The longer the effective planning horizon, the more the model’s inaccuracies introduce errors. Contrary to the conventional approach in RL, where immediate rewards are prioritized over long-term rewards via a discount factor~$\gamma \in (0,1]$, we take into account the potential impact of future risks in our methodology. Specifically, we not only scale down the mean but also scale up the variance. This informs our downstream planner about the increasing uncertainty, as the predictions extend into the future.
\begin{equation}
    \label{eq:var}
    \sigma^2_n=\sum_{t=1}^{H-1} \gamma^{-2t} \mathrm{Var[r_t^n]} + \gamma^{-2T} \mathrm{Var[\hat{R}_{T}^n]}.
\end{equation}
Accordingly, $\mathbf{U}$ is an $N \times N$ diagonal matrix 
where the $n^{\text{th}}$ element $\sigma_n$ is the  discounted standard deviation of the expected return for the $n^{\text{th}}$ trajectory.

\emph{Similarity Matrix ($\mathbf{S}$):}
In the portfolio-optimization setting, the similarity matrix typically measures the correlation between assets. In our context, the off-diagonal elements of~$\mathbf{S}$ represent the similarity between trajectories. 
To compute  element $s_{i,j}$ of $\mathbf{S}$, we use the Simple Matching Coefficient (SMC)~\cite{r:139}, which takes as input the trajectories~$\mathbf{c_i}$ and $\mathbf{c_j}$ and returns the ratio of the number of edges shared between $\mathbf{c_i}$ and $\mathbf{c_j}$ to their length (note that~$\mathbf{c_i}$ and $\mathbf{c_j}$ always share the same length).


\begin{figure}[t]
\centering
\includegraphics[clip, trim=1cm 19.1cm 9.5cm 1.5cm, width=6.5cm]{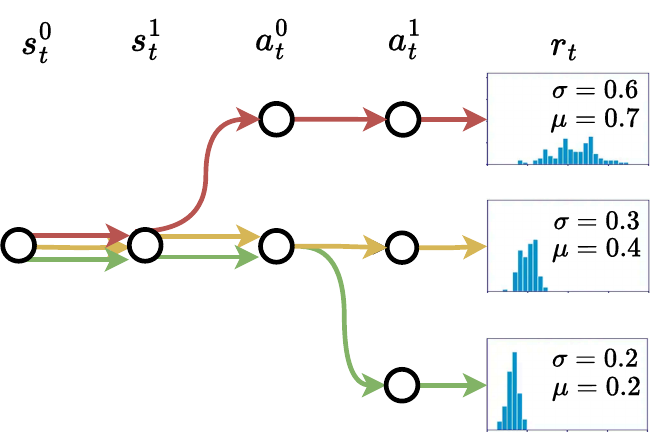} 

\caption{
\pbs tree.
}
\vspace{-1.5mm}
\label{fig:tree}
\end{figure}

\begin{example}

Fig.~\ref{fig:tree} illustrates a toy example featuring three candidate trajectories, denoted as~$\{\textcolor{myred}{\mathbf{c_1}}, \textcolor{myyellow}{\mathbf{c_2}}.
\textcolor{mygreen}{\mathbf{c_3}}\} \in\mathcal{C}$ (Alg.~\ref{alg:beam_search}). It demonstrates how to construct their corresponding matrix representations ($\bm{\mu}, \bm{\Sigma}$) within the framework of the portfolio optimization equation (Eq.~\ref{eq:portfolio_opt}). In this simplified scenario, each candidate trajectory consists of a single transition, featuring a two-dimensional state, a two-dimensional action, and a single scalar reward~$(s_t^1, s_t^2, a_t^1, a_t^2, r_t)$. Since the trajectory in this simplified case is characterized by a single reward, calculating the mean and variance of the reward using Eq.~\ref{eq:single_reward} is sufficient. However, in practice, we include all collected rewards and the reward-to-go, as shown in Eq.~\eqref{eq:var} and \eqref{eq:mean}, to construct~$\bm{\Sigma}$ and~$\bm{\mu}$.

Constructing the elements of~$\bm{\mu}$ is straightforward. Using Eq.~\ref{eq:single_reward}, we obtain $(\textcolor{myred}{\mu_1}, \textcolor{myyellow}{\mu_2}, \textcolor{mygreen}{\mu_3})=( 0.7, 0.4, 0.2 )$. To construct~$\bm{\Sigma}$, we will use the decomposition in Eq.~\ref{eq:decomposition} and represent~$\bm{\Sigma}$ with the matrices~$\mathbf{U}$ and~$\mathbf{S}$. Accordingly, using Eq.~\ref{eq:single_reward}, the elements on the diagonal of~$\mathbf{U}$ are $\textcolor{myred}{u_1}, \textcolor{myyellow}{u_2}, \textcolor{mygreen}{u_3} = (0.6, 0.3, 0.2)$. To construct the matrix~$\mathbf{S}$, we note that each trajectory has a length of four. $\textcolor{myred}{\mathbf{c_1}}$ shares one out of four edges with both~$\textcolor{myyellow}{\mathbf{c_2}}$ and~$\textcolor{mygreen}{\mathbf{c_3}}$. Using the SMC score, we find that $s_{1,2}=s_{2,1}=0.25$ and $s_{1,3}=s_{3,1}=0.25$. $\textcolor{myyellow}{\mathbf{c_2}}$ and~$\textcolor{mygreen}{\mathbf{c_3}}$ share two out of four common edges. Therefore, $s_{2,3} = s_{3,2} = 0.5$. Overall, we get:

$$
 \mathbf{S}=\begin{bmatrix}
   1 & 0.25 & 0.25 \\
   0.25 & 1 & 0.5 \\
   0.25 & 0.5 & 1
   \end{bmatrix},
 \mathbf{U}=\begin{bmatrix}
   0.6 & 0 & 0\\
   0 & 0.3 & 0\\
   0 & 0 & 0.2
   \end{bmatrix},
 \bm{\mu}=\begin{bmatrix}
   0.7  \\
   0.4 \\
   0.2 
   \end{bmatrix}.
$$
\end{example}

\subsection{{\tt \textbf{prune()}}} 
Given  coefficients $w_1, \ldots, w_N$, computed using the aforementioned {\tt \textbf{score()}} function, our {\tt \textbf{prune()}} function samples $b$ trajectories (with repetition), where the $n^{\text{th}}$ trajectory is sampled with probability $w_n$. We underline that more promising trajectories should receive greater computational investment (larger $w_n$). Consequently, these trajectories may be sampled multiple times, biasing the search towards these more promising trajectories. In such cases, our effective beam width is smaller than $b$ (but never larger).


\section{Related Work}
\label{sec:related_work}
\textbf{Offline Reinforcement Learning:}
In \emph{model-free} offline RL methods, the agent learns a policy or value function directly from the dataset. To address the distributional shift, these algorithms typically penalize the values of out-of-distribution state-action or constrain the policy closed to the behavior policy~\cite{r:17,r:21,r:22,r:82,r:67}. Additionally, uncertainty quantification techniques are used to stabilize Q-functions~\cite{r:17,r:82,r:80,r:81}. On the other side, in \emph{model-based RL} methods, the offline data is used to train predictive models of the environment that can then be used for planning or policy learning~\cite{r:92,r:24}. 
The benefits of discovering diverse solutions have been demonstrated in literature pertaining to both online and offline RL~\cite{r:140,r:141,r:142}.
However, most these methods tend to require the policy to be pessimistic, while our algorithm does not involve this constraint. Besides, our algorithm is built upon the sequence-generation framework, which is different from the above methods.

\textbf{Transformers as Trajectory Models in Offline RL:}
Decision Transformer (\dt)~\cite{r:4} predicts actions by feeding target rewards and previous states, rewards, and actions. 
Trajectory Transformer (\trt)~\cite{r:3} further brings out the capability of the sequence model by repurposing beam search as a planner. 
The Bootstrapped Trajectory Transformer (BooT)~\cite{r:134} improves on~\trt{} by self-generating additional offline data from the model to enhance its training process. The Elastic Decision Transformer~(EDT)~\cite{r:133} augments~\dt{} with a maximum in-support return training objective, enabling it to dynamically adjust history length to better combine sub-optimal trajectories. The Q-learning Decision Transformer~(QDT)~\cite{r:135} uses dynamic programming to relabel the return-to-go in the training data and trains the~\dt{} with the relabeled data. 
However, these studies primarily rely on advanced and often complex training procedures. 
In contrast, our approach focuses exclusively on modifying the inference algorithm, leaving the training process unchanged. This distinction is significant because our method is complementary and orthogonal to training-based approaches, and can potentially be combined with them to achieve additional improvements.


\textbf{Transformer Decoding Methods:}
In recent years, the rise of large Transformer-based language models has driven the development of better decoding methods. Many of these methods are sampling-based variants of \bs~\cite{r:15,r:16}. 
A notable example of diverse decoding is Diverse Beam Search~\cite{r:136}, an alternative to \bs{} that decodes a list of diverse outputs by optimizing a diversity-augmented objective. However, these methods are primarily ad-hoc and designed to produce fluent text in large Transformer-based language models trained in supervised-learning settings, where the goal is to find the most likely output sequence from the model. In offline RL, the batch also contains suboptimal behaviors, so the objective must incorporate reward signals.
\begin{table*}[h] 
    \centering
    \begin{tabular}{lrrrrrrrr}
        \toprule
        Dataset                     & DT                & QDT               & EDT               & BooT              & TT                  &\wsts (ours)               \\
        \midrule
        HalfCheetah-medium-expert   & $86.8 \pm 1.3$    & $N/A$             & $N/A$             & $94.0 \pm 1.0$    & $\mathbf{95.0} \pm 0.8$      & $92.4 \pm 9.5$   \\ 
        HalfCheetah-medium          & $42.6 \pm 0.1$    & $42.3 \pm 0.4$    & $42.5 \pm 0.9$    & $\mathbf{50.6} \pm 0.8$    & $46.9 \pm 1.6$      & $46.0 \pm 0.2$      \\ 
        HalfCheetah-medium-replay   & $36.6 \pm 0.8$    & $35.6 \pm 0.5$    & $37.8 \pm 1.5$    & $\mathbf{46.5} \pm 1.2$    & $41.9 \pm 9.7 $     & $44.5 \pm 1.3$      \\ 
        
        Hopper-medium-expert        & $107.6 \pm 1.8$   & $N/A$             & $N/A$              & $102.3 \pm 19.4$  & $110.0 \pm 10.5$    & $\mathbf{112.5} \pm 0.7$     \\ 
        Hopper-medium               & $67.6  \pm 1.0$   & $66.5 \pm 6.3$    & $63.5 \pm 5.8$    & $70.2  \pm 18.1$  & $61.1  \pm 13.9$    & $\mathbf{74.5}  \pm 6.0$     \\ 
        Hopper-medium-replay        & $82.7  \pm 7.0$   & $52.1 \pm 20.3$   & $89.0 \pm  8.3$   & $92.9  \pm 13.2$  & $91.5  \pm 13.9$    & $\mathbf{95.6}  \pm 9.7$     \\ 
        
        Walker2d-medium-expert      & $108.1 \pm 0.2$   & $N/A$             & $N/A$              & $\mathbf{110.4} \pm 2.0$   & $101.9 \pm 26.3$    & $108.9 \pm 0.8$     \\ 
        Walker2d-medium             & $74.0 \pm 1.4$    & $67.1 \pm 3.2$    & $72.8 \pm 6.2$    & $\mathbf{82.9}  \pm 11.7$  & $79.0  \pm 10.8$    & $79.8  \pm 5.4$     \\ 
        Walker2d-medium-replay      & $66.6 \pm 3.0$    & $58.2 \pm 5.1$    & $74.8 \pm 4.9$    & $87.6  \pm 9.9$   & $82.6  \pm 26.7$    & $\mathbf{87.6}  \pm 5.7$     \\ 

        \midrule
        Average Return              & $74.7 (61.7)$     & $N/A (53.6)$      & $N/A (63.4)$        & $81.9 (71.8)$       & $78.9 (67.2) $        & $\mathbf{82.4} (71.3)$          \\
        Average Standard Deviation  & $1.84 (2.22)$     & $N/A (5.97)$      & $N/A (4.6)$         & $8.59 (9.15)$       & $12.7 (12.8)$         & $\mathbf{4.4} (4.7)  $            \\

        \bottomrule
    \end{tabular}
    \caption{The table presents baseline comparisons on D4RL tasks. Our results are evaluated based on 30 random evaluations. The results for other baselines are adopted from their reported scores. For \trt{}, the paper reported the standard error; however, as we are more interested in the standard deviation, we converted their results accordingly. 
    Since QDT and EDT do not provide results for the medium expert dataset, we have excluded these datasets and listed the results of the other algorithms in parentheses for comparison. All tests were conducted on NVIDIA GeForce RTX 3090 GPUs.}
    \label{tab:results}
\end{table*}

\section{Experiments and Results}

To evaluate \pbs, we empirically compare it with three methods based on the Trajectory Transformer:
the original Trajectory Transformer (\trt)~\cite{r:3}.
the Bootstrapped Trajectory Transformer (BooT)~\cite{r:134} and 
Elastic Decision Transformer (EDT)~\cite{r:133}.
Additionally, we add comparisons with Decision Transformer (\dt)~\cite{r:4} and an extension of the Q-learning Decision Transformer (QDT)~\cite{r:135} 

We evaluate our algorithm on locomotion control tasks using the OpenAI Gym MuJoCo physics engine~\cite{r:127}. To ensure that our performance improvements are solely due to the improved Transformer-decoding algorithms, we used the same hyperparameters (including beam width, planning horizon, vocabulary size, and context size) and trained models with identical network weights as those provided in the original~\trt, which were trained on D4RL offline dataset for locomotion tasks~\cite{r:9}. 

D4RL consists of various datasets designed for benchmarking offline RL. The ``medium'' dataset comes from a policy reaching about a third of expert performance.
The ``medium-replay'', sourced from this policy’s replay buffer, and the ``medium-expert'' dataset is generated by mixing samples generated by the medium policy and samples generated by an expert policy.

We present Gaussian distributions illustrating the mean and variance of the baseline algorithms results in Fig.~\ref{fig:mujoco}. Detailed results can be found in the Appendix, Table~\ref{tab:results}. The results indicate that the proposed~\wsts{} consistently outperforms the baseline methods~\trt{},~\dt{}, and their variants across the majority of datasets. Notably,~\wsts{} reduces the standard deviation by more than ${60\%}$ compared to the \trt{} network, which shares the same weights, while also achieving a higher mean. BooT achieved higher means on some benchmarks; however, its overall average score is lower, and its average standard deviation is more than twice as high.
These findings highlight the effectiveness of our approach in addressing distribution shift. Consequently, \wsts{} demonstrates more stable behavior, which is crucial in high-stakes offline RL settings where errors can have significant consequences. 

\label{sec:result}
\begin{figure}[t]
\centering
\includegraphics[width=8.6cm]{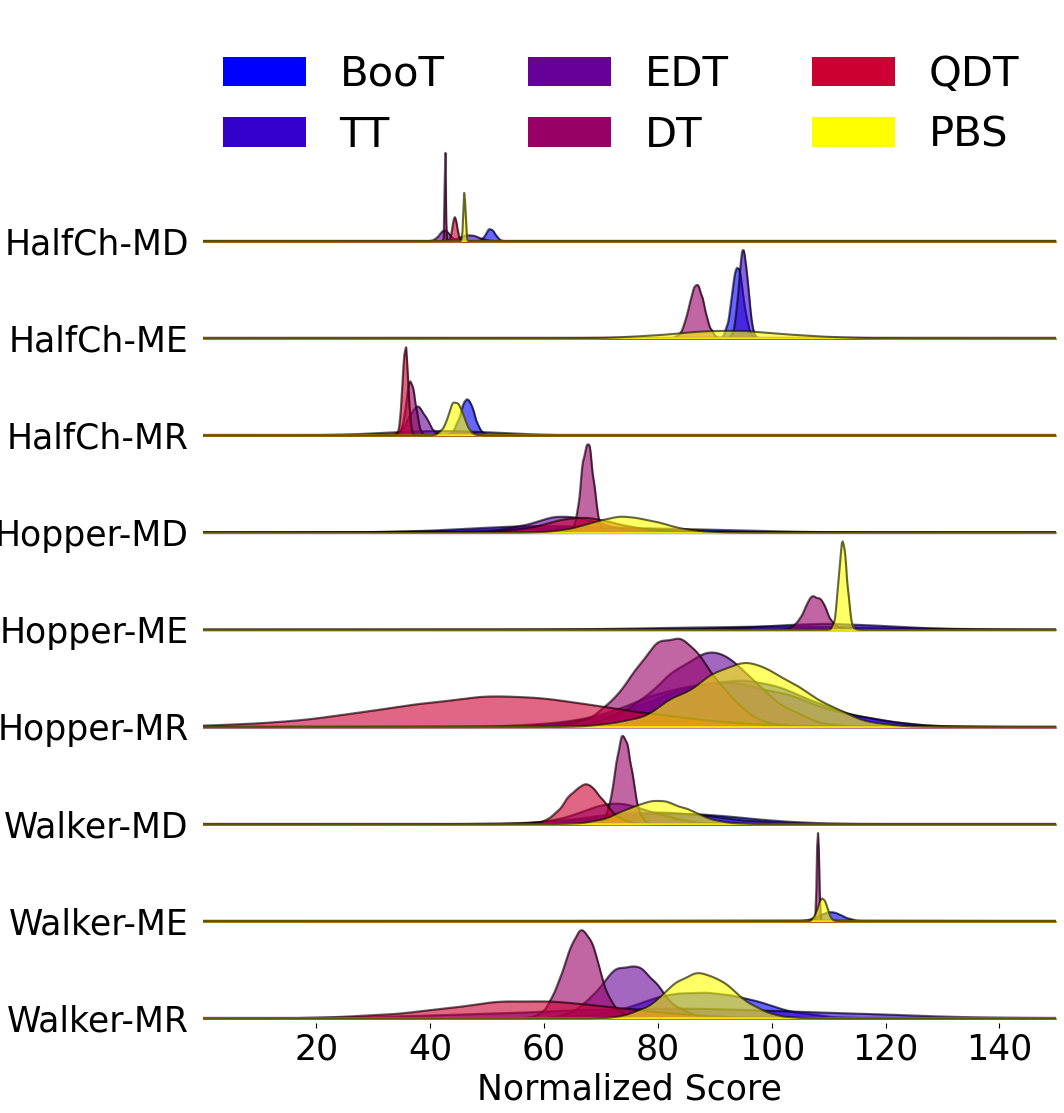} 
\caption{
Scores of the Gym-MuJoCo tasks: HalfCheetah, Hopper, and Walker2d, trained on medium (MD), medium-replay (MR), and medium-expert (ME) datasets.
}
\label{fig:mujoco}
\end{figure}

For all datasets and environments in our reported results (Table~\ref{tab:results}), except for HalfCheetah-medium-expert, we set $\delta=1$, giving equal weights to the expected return~($\bm{\mu}$) and the covariance matrix~($\bm{\Sigma}$) (terms (1) and (2) in Eq.~\eqref{eq:portfolio_opt}). Additionally, we set $\alpha=0.1$ and the discount factor $\gamma=0.99$. HalfCheetah-medium-expert is the only dataset where our performance is worse than the original \bs{} variant, which achieves a remarkably low standard deviation. Interestingly, adopting a more aggressive approach can be advantageous in this case. We observed improved results for more risk-tolerant settings ($\alpha=\delta=0.01$). 
This risk-tolerant setting results in numerous negligible weights in the portfolio, leading to the exclusion of most trajectories from the beam. Consequently, this creates an almost greedy decoding with small beam width ($\vert \mathcal{C} \vert \leq 10$) for most time steps. However, for this dataset, our performance still lags the original expectation-maximizing baseline. Although further adapting these parameters for each dataset could potentially enhance performance, we opted not to do so, as our algorithm is robust enough to demonstrate improvements without additional tuning.

The results underscore the pivotal role of the decoding algorithm in determining the quality of the generated sequence in offline RL. This observation is consistent with findings in natural language processing, where the choice of decoding strategies can significantly impact the quality of machine-generated text, even when using the same neural language model~\cite{r:15}.

\section{Conclusion}
\label{sec:conclusion}
In this work, we addressed the question of how we can modify the decoding algorithms to tackle extremely challenging offline RL problems such as locomotion problems.
To this end, we introduced \wsts, a decoding algorithm that adapts tools from a different domain (economics) and proposes a structured approach to risk control, pinpointing where these tools can be effectively utilized in the decoding process.

\wsts{} leverages the model's ability to generalize to states outside the static dataset support. However, it remains cautious when drifting to states where the model cannot provide confident predictions based on the dataset. Additionally it encourage diversity which is important in problems where multiple solutions exist. This dual consideration allows for a robust selection of trajectories, a crucial attribute in offline RL, where failures can lead to significant costs or dangerous consequences. 
While many studies focus on enhancing Transformer training procedures, we emphasize that \wsts{} is exclusively an inference-time algorithm. Consequently, it operates independently of these methods and can be integrated with advanced training techniques to further boost performance.





\newpage
\bibliographystyle{named}
\bibliography{ijcai25}

\begin{thebibliography}{}

\bibitem[\protect\citeauthoryear{Agarwal \bgroup \em et al.\egroup }{2020}]{r:80}
Rishabh Agarwal, Dale Schuurmans, and Mohammad Norouzi.
\newblock An optimistic perspective on offline reinforcement learning.
\newblock In {\em {International Conference on Machine Learning (ICML)}}, page 104–114, 2020.

\bibitem[\protect\citeauthoryear{Argenson and Dulac-Arnold}{2020}]{r:92}
Arthur Argenson and Gabriel Dulac-Arnold.
\newblock Model-based offline planning.
\newblock In {\em {International Conference on Learning Representations (ICLR)}}, pages 14129--14142, 2020.

\bibitem[\protect\citeauthoryear{Chen \bgroup \em et al.\egroup }{2021}]{r:4}
Lili Chen, Kevin Lu, Aravind Rajeswaran, Kimin Lee, Aditya Grover, Michael Laskin, Pieter Abbeel, Aravind Srinivas, and Igor Mordatch.
\newblock Decision transformer: Reinforcement learning via sequence modeling.
\newblock In {\em {Advances in Neural Information Processing Systems (NeurIPS)}}, pages 15084 -- 15097, 2021.

\bibitem[\protect\citeauthoryear{Covington \bgroup \em et al.\egroup }{2016}]{r:78}
Paul Covington, Jay Adams, and Emre Sargin.
\newblock Deep neural networks for youtube recommendations.
\newblock In {\em ACM Conference on Recommender Systems}, pages 191--198, 2016.

\bibitem[\protect\citeauthoryear{Desai and Durrett}{2020}]{r:45}
Shrey Desai and Greg Durrett.
\newblock Calibration of pre-trained transformers.
\newblock In {\em {Empirical Methods in Natural Language Processing (EMNLP)}}, pages 295--302, 2020.

\bibitem[\protect\citeauthoryear{Fan \bgroup \em et al.\egroup }{2018}]{r:16}
Angela Fan, Mike Lewis, and Yann Dauphin.
\newblock Hierarchical neural story generation.
\newblock In {\em {Association for Computational Linguistics (ACL)}}, pages 889--898, 2018.

\bibitem[\protect\citeauthoryear{Fu \bgroup \em et al.\egroup }{2020}]{r:9}
Justin Fu, Aviral Kumar, Ofir Nachum, George Tucker, and Sergey Levine.
\newblock D4rl: Datasets for deep data-driven reinforcement learning, 2020.

\bibitem[\protect\citeauthoryear{Fujimoto \bgroup \em et al.\egroup }{2019}]{r:17}
Scott Fujimoto, David Meger, and Doina Precup.
\newblock Off-policy deep reinforcement learning without exploration.
\newblock In {\em {International Conference on Machine Learning (ICML)}}, pages 2052--2062, 2019.

\bibitem[\protect\citeauthoryear{Goodman}{2018}]{r:61}
Joseph Goodman.
\newblock {\em The Best Quotes Book: 555 Daily Inspirational and Motivational Quotations by Famous People}.
\newblock Lulu, 2018.

\bibitem[\protect\citeauthoryear{Gottesman \bgroup \em et al.\egroup }{2018}]{r:63}
Omer Gottesman, Fredrik Johansson, Joshua Meier, Jack Dent, Donghun Lee, Srivatsan Srinivasan, Linying Zhang, Yi~Ding, David Wihl, Xuefeng Peng, Jiayu Yao, Isaac Lage, Christopher Mosch, Li-wei~H. Lehman, Matthieu Komorowski, Matthieu Komorowski, Aldo Faisal, Leo~Anthony Celi, David Sontag, and Finale Doshi-Velez.
\newblock Evaluating reinforcement learning algorithms in observational health settings, 2018.

\bibitem[\protect\citeauthoryear{Guo \bgroup \em et al.\egroup }{2017}]{r:46}
Chuan Guo, Geoff Pleiss, Yu~Sun, and Kilian~Q. Weinberger.
\newblock On calibration of modern neural networks.
\newblock In {\em {International Conference on Machine Learning (ICML)}}, pages 1321--1330, 2017.

\bibitem[\protect\citeauthoryear{Holtzman \bgroup \em et al.\egroup }{2019}]{r:15}
Ari Holtzman, Jan Buys, Leo Du, Maxwell Forbes, and Yejin Choi.
\newblock The curious case of neural text degeneration.
\newblock In {\em CEUR Workshop Proceedings}, volume 2540, 2019.

\bibitem[\protect\citeauthoryear{Janner \bgroup \em et al.\egroup }{2021}]{r:3}
Michael Janner, Qiyang Li, and Sergey Levine.
\newblock Offline reinforcement learning as one big sequence modeling problem.
\newblock In {\em {Advances in Neural Information Processing Systems (NeurIPS)}}, pages 1273 -- 1286, 2021.

\bibitem[\protect\citeauthoryear{Jaques \bgroup \em et al.\egroup }{2019}]{r:81}
Natasha Jaques, Asma Ghandeharioun, Judy~Hanwen Shen, Craig Ferguson, Agata Lapedriza, Noah Jones, Shixiang Gu, and Rosalind Picard.
\newblock Way off-policy batch deep reinforcement learning of implicit human preferences in dialog, 2019.

\bibitem[\protect\citeauthoryear{Kidambi \bgroup \em et al.\egroup }{2020}]{r:24}
Rahul Kidambi, Aravind Rajeswaran, Praneeth Netrapalli, and Thorsten Joachims.
\newblock Morel : Model-based offline reinforcement learning.
\newblock In {\em {Advances in Neural Information Processing Systems (NeurIPS)}}, pages 21810--21823, 2020.

\bibitem[\protect\citeauthoryear{Kumar \bgroup \em et al.\egroup }{2019}]{r:21}
Aviral Kumar, Justin Fu, George Tucker, and Sergey Levine.
\newblock Stabilizing off-policy q-learning via bootstrapping error reduction.
\newblock In {\em {Advances in Neural Information Processing Systems (NeurIPS)}}, pages 11761--11771, 2019.

\bibitem[\protect\citeauthoryear{Kumar \bgroup \em et al.\egroup }{2020a}]{r:23}
Aviral Kumar, Aurick Zhou, George Tucker, and Sergey Levine.
\newblock Conservative q-learning for offline reinforcement learning.
\newblock In {\em {Advances in Neural Information Processing Systems (NeurIPS)}}, pages 1179--1191, 2020.

\bibitem[\protect\citeauthoryear{Kumar \bgroup \em et al.\egroup }{2020b}]{r:142}
Aviral Kumar, Aurick Zhou, George Tucker, and Sergey Levine.
\newblock Conservative q-learning for offline reinforcement learning.
\newblock In {\em {Advances in Neural Information Processing Systems (NeurIPS)}}, pages 1179 -- 1191, 2020.

\bibitem[\protect\citeauthoryear{Kumar \bgroup \em et al.\egroup }{2020c}]{r:140}
Saurabh Kumar, Aviral Kumar, Sergey Levine, and Chelsea Finn.
\newblock One solution is not all you need: Few-shot extrapolation via structured maxent rl.
\newblock In {\em {Advances in Neural Information Processing Systems (NeurIPS)}}, pages 8198 -- 8210, 2020.

\bibitem[\protect\citeauthoryear{Kumar \bgroup \em et al.\egroup }{2020d}]{r:132}
Shakti Kumar, Jerrod Parker, and Panteha Naderian.
\newblock Adaptive transformers in rl.
\newblock {\em ArXiv}, 2004.03761, 2020.

\bibitem[\protect\citeauthoryear{Li \bgroup \em et al.\egroup }{2023}]{r:138}
Wenzhe Li, Hao Luo, Zichuan Lin, Chongjie Zhang, Zongqing Lu, and Deheng Ye.
\newblock A survey on transformers in reinforcement learning.
\newblock {\em Transactions on Machine Learning Research}, 2023.

\bibitem[\protect\citeauthoryear{Liu \bgroup \em et al.\egroup }{2023}]{r:124}
Haotian Liu, Chunyuan Li, Qingyang Wu, and Yong~Jae Lee.
\newblock Visual instruction tuning.
\newblock In {\em {Advances in Neural Information Processing Systems (NeurIPS)}}, 2023.

\bibitem[\protect\citeauthoryear{Markowitz}{1952}]{r:6}
Harry Markowitz.
\newblock Portfolio selection.
\newblock {\em The Journal of Finance}, 7(1):77--91, 1952.

\bibitem[\protect\citeauthoryear{Murphy}{1996}]{r:139}
Allan~H. Murphy.
\newblock The finley affair: A signal event in the history of forecast verification.
\newblock {\em Weather and Forecasting}, 11:3--20, 1996.

\bibitem[\protect\citeauthoryear{Osa and Harada}{2024}]{r:141}
Takayuki Osa and Tatsuya Harada.
\newblock Discovering multiple solutions from a single task in offline reinforcement learning.
\newblock In {\em {International Conference on Machine Learning (ICML)}}, pages 38864 -- 38884, 2024.

\bibitem[\protect\citeauthoryear{Parisotto and Salakhutdinov}{2021}]{r:131}
Emilio Parisotto and Russ Salakhutdinov.
\newblock Efficient transformers in reinforcement learning using actor-learner distillation.
\newblock In {\em {International Conference on Learning Representations (ICLR)}}, 2021.

\bibitem[\protect\citeauthoryear{Peng \bgroup \em et al.\egroup }{2019}]{r:67}
Xue~Bin Peng, Aviral Kumar, Grace Zhang, and Sergey Levine.
\newblock Advantage-weighted regression: Simple and scalable off-policy reinforcement learning, 2019.

\bibitem[\protect\citeauthoryear{Radford \bgroup \em et al.\egroup }{2023}]{r:122}
Alec Radford, Jong~Wook Kim, Tao Xu, Greg Brockman, Christine McLeavey, and Ilya Sutskever.
\newblock Robust speech recognition via large-scale weak supervision.
\newblock In {\em {International Conference on Machine Learning (ICML)}}, pages 7487--7498, 2023.

\bibitem[\protect\citeauthoryear{Reed \bgroup \em et al.\egroup }{2022}]{r:129}
Scott Reed, Konrad Zolna, Emilio Parisotto, Sergio~G{\'o}mez Colmenarejo, Alexander Novikov, Gabriel Barth-maron, Mai Gim{\'e}nez, Yury Sulsky, Jackie Kay, Jost~Tobias Springenberg, Tom Eccles, Jake Bruce, Ali Razavi, Ashley Edwards, Nicolas Heess, Yutian Chen, Raia Hadsell, Oriol Vinyals, Mahyar Bordbar, and Nando de~Freitas.
\newblock A generalist agent.
\newblock {\em Transactions on Machine Learning Research}, 2022.

\bibitem[\protect\citeauthoryear{Sallab \bgroup \em et al.\egroup }{2017}]{r:79}
Ahmad~EL Sallab, Mohammed Abdou, Etienne Perot, and Senthil Yogamani.
\newblock Deep reinforcement learning framework for autonomous driving.
\newblock {\em Electronic Imaging}, 19:70--76, 2017.

\bibitem[\protect\citeauthoryear{Siegel \bgroup \em et al.\egroup }{2020}]{r:22}
Noah~Y Siegel, Jost~Tobias Springenberg, Felix Berkenkamp, Michael Neunert, Thomas Lampe, Roland Hafner, and Martin Riedmiller.
\newblock Keep doing what worked : Behavioral cloning priors for fully offline learning.
\newblock In {\em {International Conference on Learning Representations (ICLR)}}, 2020.

\bibitem[\protect\citeauthoryear{Strehl \bgroup \em et al.\egroup }{2010}]{r:77}
Alex Strehl, John Langford, Lihong Li, and Sham~M Kakade.
\newblock Learning from logged implicit exploration data.
\newblock In {\em {Advances in Neural Information Processing Systems (NIPS)}}, pages 217--–2225, 2010.

\bibitem[\protect\citeauthoryear{Sutton and Barto}{1998}]{r:1}
R.S. Sutton and A.G. Barto.
\newblock {\em Reinforcement Learning: An Introduction}.
\newblock {MIT} Press, 1998.

\bibitem[\protect\citeauthoryear{Todorov \bgroup \em et al.\egroup }{2012}]{r:127}
Emanuel Todorov, Tom Erez, and Yuval Tassa.
\newblock Mujoco: a physics engine for model-based control.
\newblock In {\em {International Conference on Intelligent Robots and Systems (IROS)}}, pages 5026--5033, 2012.

\bibitem[\protect\citeauthoryear{Vaswani \bgroup \em et al.\egroup }{2017}]{r:57}
Ashish Vaswani, Noam Shazeer, Niki Parmar, Jakob Uszkoreit, Llion Jones, Aidan~N. Gomez, Lukasz Kaiser, and Illia Polosukhin.
\newblock Attention is all you need.
\newblock In {\em {Advances in Neural Information Processing Systems (NIPS)}}, page 6000–6010, 2017.

\bibitem[\protect\citeauthoryear{Vijayakumar \bgroup \em et al.\egroup }{2018}]{r:136}
Ashwin~K Vijayakumar, Michael Cogswell, Ramprasath~R. Selvaraju, Qing Sun, Stefan Lee, David Crandall, and Dhruv Batra.
\newblock Diverse beam search: Decoding diverse solutions from neural sequence models, 2018.

\bibitem[\protect\citeauthoryear{Wang \bgroup \em et al.\egroup }{2018}]{r:75}
Lu~Wang, Wei Zhang, Xiaofeng He, and Hongyuan Zha.
\newblock Supervised reinforcement learning with recurrent neural network for dynamic treatment recommendation.
\newblock In {\em {International Conference on Knowledge Discovery and Data Mining (SIGKDD)}}, pages 2447--2456, 2018.

\bibitem[\protect\citeauthoryear{Wang \bgroup \em et al.\egroup }{2022}]{r:134}
Kerong Wang, Hanye Zhao, Xufang Luo, Kan Ren, Weinan Zhang, and Dongsheng Li.
\newblock Bootstrapped transformer for offline reinforcement learning.
\newblock In {\em {Advances in Neural Information Processing Systems (NeurIPS)}}, 2022.

\bibitem[\protect\citeauthoryear{Wu \bgroup \em et al.\egroup }{2019}]{r:82}
Yifan Wu, George Tucker, and Ofir Nachum.
\newblock Behavior regularized offline reinforcement learning, 2019.

\bibitem[\protect\citeauthoryear{Wu \bgroup \em et al.\egroup }{2023}]{r:133}
Yueh-Hua Wu, Xiaolong Wang, and Masashi Hamaya.
\newblock Elastic decision transformer.
\newblock In {\em {Advances in Neural Information Processing Systems (NeurIPS)}}, pages 18532 -- 18550, 2023.

\bibitem[\protect\citeauthoryear{Wu \bgroup \em et al.\egroup }{2024}]{r:130}
Yueh-Hua Wu, Xiaolong Wang, and Masashi Hamaya.
\newblock Elastic decision transformer.
\newblock In {\em {Advances in Neural Information Processing Systems (NeurIPS)}}, pages 18532 -- 18550, 2024.

\bibitem[\protect\citeauthoryear{Yamagata \bgroup \em et al.\egroup }{2023}]{r:135}
Taku Yamagata, Ahmed Khalil, and Ra\'{u}l Santos-Rodr\'{\i}guez.
\newblock Q-learning decision transformer: Leveraging dynamic programming for conditional sequence modeling in offline rl.
\newblock In {\em {International Conference on Machine Learning (ICML)}}, pages 38989 -- 39007, 2023.

\bibitem[\protect\citeauthoryear{Yao \bgroup \em et al.\egroup }{2023}]{r:128}
Shunyu Yao, Jeffrey Zhao, Dian Yu, Nan Du, Izhak Shafran, Karthik~R. Narasimhan, and Yuan Cao.
\newblock React: Synergizing reasoning and acting in language models.
\newblock In {\em {International Conference on Learning Representations (ICLR)}}, 2023.

\bibitem[\protect\citeauthoryear{Yu \bgroup \em et al.\egroup }{2019}]{r:76}
Chao Yu, Guoqi Ren, and Jiming Liu.
\newblock Deep inverse reinforcement learning for sepsis treatment.
\newblock In {\em International Conference on Healthcare Informatics (ICHI)}, pages 1--3, 2019.

\end{thebibliography}

\end{document}